\documentclass[runningheads]{llncs}
\usepackage{graphicx}
\usepackage{bm}
\usepackage{amsfonts,amssymb}
\usepackage{multicol}
\usepackage{multirow}
\begin{document}
%
    \title{Mixed-Supervised Dual-Network for Medical Image Segmentation}
\titlerunning{Mixed-Supervised Dual-Network}
%

\author{Duo Wang\inst{1,2} \and
Ming Li\inst{3} \and Nir Ben-Shlomo\inst{4} \and C. Eduardo
Corrales\inst{4,5} \and Yu Cheng\inst{6} \and Tao Zhang\inst{1} \and
Jagadeesan Jayender\inst{2,5}}



%
%

\institute{Department of Automation, Tsinghua University, Beijing,
China \and
    Department of Radiology, Brigham and Women's Hospital, Boston, USA \and
Department of Radiology and Radiation Oncology, Huadong Hospital
affiliated to Fudan University, Shanghai, China \and Department of
Surgery, Brigham and Women's Hospital, Boston, USA \and Harvard
Medical School, Boston, USA \and Microsoft AI \& Research, Redmond,
Washington, USA}

%
\maketitle              
\begin{abstract}
Deep learning based medical image segmentation models usually
require large datasets with high-quality dense segmentations to
train, which are very time-consuming and expensive to prepare. One
way to tackle this challenge is using the mixed-supervised learning
framework, in which only a part of data is densely annotated with
segmentation label and the rest is weakly labeled with bounding
boxes. The model is trained jointly in a multi-task learning
setting.
In this paper, we propose Mixed-Supervised Dual-Network (MSDN), a
novel architecture which consists of two separate networks for the
detection and segmentation tasks respectively, and a series of
connection modules between the layers of the two networks. These
connection modules are used to transfer useful information from the
auxiliary detection task to help the segmentation task. We propose
to use a recent technique called 'Squeeze and Excitation' in the
connection module to boost the transfer. We conduct experiments on
two medical image segmentation datasets. The proposed MSDN model
outperforms multiple baselines.

\end{abstract}
\section{Introduction}
Image segmentation is an important application of medical image
analysis. Recently, deep learning based
methods~\cite{ref2,ref3,ref7,ref_add} have achieved remarkable
success in many medical image segmentation tasks, such as brain
tumor and lung nodule segmentation. However, all these methods
require a large amount of training data with high-quality dense
annotations to train, which is very expensive and time-consuming to
prepare.

Therefore, weakly-supervised segmentation with insufficient labels,
e.g. image tags~\cite{ref8} or bounding boxes~\cite{ref9} has
attracted a lot of attention recently. Although great progress has
been made, there still exists some gap in performance compared to
the models trained with fully-supervised datasets. This makes it
impractical for the medical image scenario, where accurate
segmentation maps are required for disease diagnosis, surgical
planning or pathological analysis. On the other hand, these
weakly-supervised models are usually trained in multi-step iteration
mode~\cite{ref8,ref9} or with prior medical knowledge~\cite{ref11},
making it difficult to be scalable on real applications.

Another promising approach is the mixed-supervised segmentation,
where only a part of data is densely annotated with segmentation map
and the rest is labeled with weak form (such as with bounding
boxes). Typical existing methods~\cite{ref12,ref18,ref19} consider
training with such kind of data in a multi-task learning setting and
exploit multi-stream network, where basic feature extractor is
shared and different streams are used for data with different
annotation forms. The work in~\cite{ref19} focuses on the optimal
balance between the number of annotations needed for different
supervision types and presents a budget-based cost-minimization
framework in a mixed-supervision setting.

In this paper, we propose a novel architecture for mixed-supervised
medical image segmentation. Considering the bounding boxes as weak
annotation, our method takes the segmentation task as target task,
which is augmented with object detection task (auxiliary task).
Different from the multi-stream structure with shared
backbone~\cite{ref12}, our new architecture is made up of two
separate networks for each task. The two networks are linked by a
series of connection modules that lie between the corresponding
layers. These connection modules take as input the convolution
features of detection network and transfer useful information to the
segmentation network to help the training of the segmentation task.
We propose to use a recent feature attention technique called
``Squeeze and Excitation"~\cite{ref13,ref14,ref17} in the connection
module to boost the information transfer. The proposed model is
named as Mixed-Supervised Dual-Network (MSDN). We perform evaluation
on the lung nodule segmentation and the cochlea segmentation of CT
images. Experimental results show that our model is able to
outperform multiple baseline approach in both datasets.

\section{Methods}
\subsection{Squeeze and Excitation}
\begin{figure}
\includegraphics[width=\textwidth]{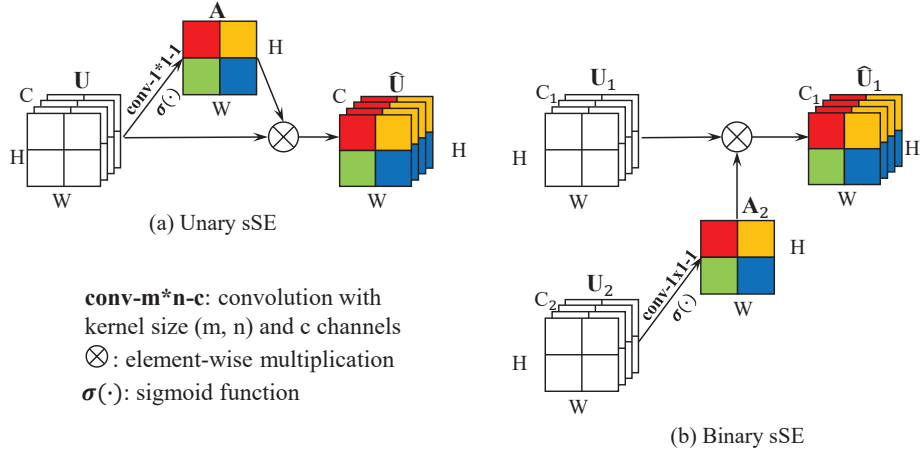}
\caption{Illustration of the Channel Squeeze and Spatial Excitation
(sSE) architecture of Unary form (a) and Binary form (b).}
\label{fig1}
\end{figure}

``Squeeze-and-Excitation" (SE) was first introduced in \cite{ref13}
and can be flexibly integrated in any CNN model. The SE module first
squeezes the feature map by global average pooling and then passes
the squeezed feature to the gating module to get the representation
of channel-wise dependencies, which is used to re-calibrate the
feature map to emphasize on useful channels. The work
in~\cite{ref14} refers to the SE module in \cite{ref13} as Spatial
Squeeze and Channel Excitation (cSE) and proposes a different
version called Channel Squeeze and Spatial Excitation (sSE). The sSE
module squeezes the feature map along channels to preserve more
spatial information, thus is more suitable for image segmentation
task. The two SE modules mentioned above are unary, as both the
squeeze and excitation are operated on the same feature map. Abhijit
\textit{et al.}~\cite{ref17} builds a binary version of sSE and
applies it to their two-armed architecture for few-shot
segmentation. Since the sSE module is related to our method, we will
give a more detailed introduction as follows (see Fig.~\ref{fig1}).

We consider the feature map \textbf{U} = [$\textbf{\textrm{u}}^{1}$,
$\textbf{\textrm{u}}^{2}$, ..., $\textbf{\textrm{u}}^{C}$] from
previous convolution layer as the input of the Unary sSE module and
\bm{$u^{i}$} $\in$ $\mathbb{R}^{W \times H}$ denotes its $i$th
channel. The channel squeeze operation is achieved by 1 $\times$ 1
convolution with kernel weight $\textbf{\textrm{w}}_{sq}$ $\in$
$\mathbb{R}^{1 \times C \times 1 \times 1}$. The squeezed feature is
then passed through sigmoid function to derive the attention weight
\textbf{A} $\in$ $\mathbb{R}^{W \times H}$. Then each feature
channel of \textbf{U} is multiplied element-wise by \textbf{A} to
get the spatially recalibrated feature
$\widehat{\textbf{\textrm{U}}}$ as output
\begin{equation}
\widehat{\mathbf{U}} =
[\sigma(\mathbf{w}_{sq}*\mathbf{U})\circ\mathbf{u}^{1},
\sigma(\mathbf{w}_{sq}*\mathbf{U})\circ\mathbf{u}^{2}, ...,
\sigma(\mathbf{w}_{sq}*\mathbf{U})\circ\mathbf{u}^{C}]
\end{equation}
Here $\circ$ denotes the Hadamard product, * denotes the convolution
operation and $\sigma$ denotes the sigmoid function.

Binary sSE extends the idea of Unary sSE, which takes two feature
maps as inputs. One feature map is squeezed and used to recalibrate
the other feature as output
\begin{equation}
\widehat{\mathbf{U}}_{1} =
[\sigma(\mathbf{w}_{sq}*\mathbf{U}_{2})\circ\mathbf{u}^{1}_{1},
\sigma(\mathbf{w}_{sq}*\mathbf{U}_{2})\circ\mathbf{u}^{2}_{1}, ...,
\sigma(\mathbf{w}_{sq}*\mathbf{U}_{2})\circ\mathbf{u}^{C}_{1}]
\end{equation}
We propose to use the Binary sSE module as the connection between
our dual-network architecture for information extraction and
transfer.

\subsection{Architectural Design}
\begin{figure}
    \includegraphics[width=\textwidth]{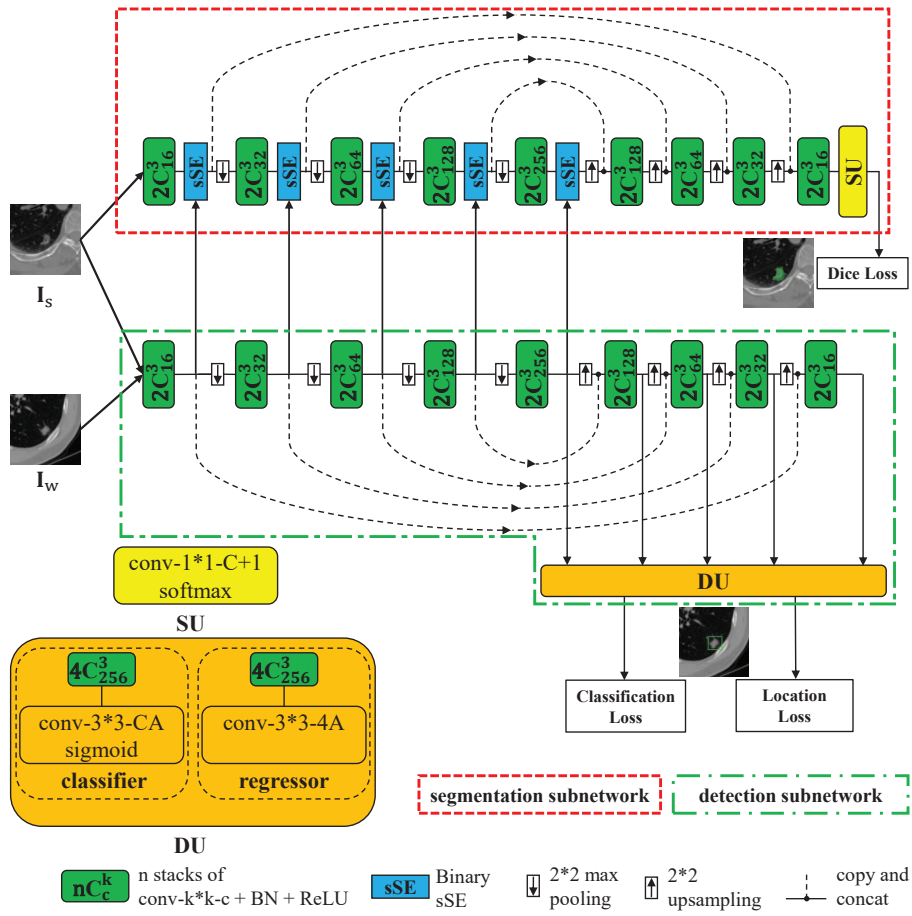}
    \caption{Structure of Mixed-Supervised Dual-Network (MSDN).} \label{fig2}
\end{figure}

Our MSDN follows the setting of multi-task learning and is made up
of two separate sub-networks for the segmentation and detection
tasks respectively (as shown in Fig.~\ref{fig2}). Both sub-networks
are built from the U-Net and contain 9 feature stages, with 4 stages
in the Encoder, 4 in the Decoder and 1 in the Bottleneck.  Each
feature stage consists of 2 dilated-convolution layers with 3*3
kernel, each followed by batch normalization~\cite{ref16} and
rectified linear unit(ReLU). The output of each Encoder stage is
skip-connected to the corresponding Decoder stage to recover spatial
information lost during max-pooling \cite{ref24}. Dilation factors
are set as [1,2,2,2,4,2,2,2,1] in the 9 feature stages respectively.
The stride and padding are chosen accordingly to make the size of
the output feature identical to that of the input.
For the segmentation sub-network, the sSE modules are added after
each stage in its Encoder and Bottleneck, which take the
segmentation and detection features from the same stage as input,
squeeze the detection feature and re-calibrate the segmentation
feature. In this way, the segmentation sub-network can extract
useful information from the auxiliary detection sub-network to
facilitate its training.

The Segmentation Unit (\textbf{SU}) takes the extracted features
into a 1*1 convolution layer followed by a channel-wise soft-max to
output a dense segmentation map with C+1 channels, where C is the
number of segmentation classes and we treat the background as
another class. Dice loss~\cite{ref3} is used for the segmentation
sub-network training.

For the detection sub-network, we build the Detection Unit
(\textbf{DU}) under a single-stage object-detection paradigm,
similarly to \cite{ref12,ref15}. The \textbf{DU} consists of a
classifier block and a bounding box regressor block and takes as
input the convolution feature from the detection sub-network and
produces class predictions for C target classes and object locations
via bounding boxes. Note that all features from the Decoder stages
are used for detection and the parameters of \textbf{DU} are shared.
At each position of the feature, totally A=9 reference bounding
boxes of different shapes and sizes are built as anchors. The
\textbf{DU} predicts the class label (C-length vector) of the object
and the relative position (4-length vector) to the near ground-truth
bounding boxes for each of the A anchors. Thus, the
classifier(regressor) takes the feature from the Decoder of the
detection sub-network through 4 3*3 convolution layers with 256
channels and one 3*3 convolution layer with C*A(4*A) channels. A
Sigmoid function is used to scale the output of classifier to [0,
1]. The detection loss is the sum of the cross-entropy based focal
loss for classification and the regularized-L1 loss for
location~\cite{ref15}.

During training, we mix the strongly- and weakly-annotated data and
shuffle them. At each training iteration, we randomly select a batch
of data as the input to our model. The strongly-annotated data
$\textbf{\textrm{I}}_\textrm{s}$ goes through the Encoder of the
detection and segmentation sub-networks and its segmentation
features are re-calibrated by the detection features for the Decoder
to derive the segmentation loss. The weakly-annotated data
$\textbf{\textrm{I}}_\textrm{w}$ only goes through the detection
sub-network to get the detection loss. The sum of the segmentation
loss and detection loss is minimized to train the model.


The structure of our model is similar to that in \cite{ref17}, as we
both build a dual architecture with two sub-networks and exploit sSE
modules as connection. However, the work~\cite{ref17} focuses on the
few-shot segmentation problem. The two sub-networks are used for the
same segmentation task and trained jointly in the meta-learning
mode. sSE modules exist in every feature stage of the base network.
While, our model is designed for mixed-supervised segmentation
problem, the two networks are used for different tasks and trained
iteratively in multi-task learning mode. Because of that, the
features of the two sub-networks in shallow layers may be relative
to each other and those in deep layers may be task-specific. So we
only use sSE in the shallow layers, specifically, in the Encoder and
Bottleneck.

\section{Experiments}
We evaluate our model on two medical image segmentation datasets:
lung nodule dataset and cochlea of inner ear dataset. The lung
dataset consists of non-contrast CT of 320 nodules that was acquired
on a 64-detector CT system (GE Light Speed VCT or GE Discovery CT750
HD, GE Healthcare, Milwaukee, WI, USA) using the scan parameters:
section width, 1.25 mm; reconstruction interval, 1.25 mm; pitch,
0.984; 120 kV; and 35 mA; display field of view (DFOV) ranged from
28cm to 36 cm; matrix size, 512*512, pixel size ranged from 0.55mm
to 0.7mm. We randomly choose 160, 80 and 80 as training, validating
and testing dataset. The inner ear dataset consisted of non-contrast
temporal bone CT of 146 cochleas that was acquired on a Siemens
Somatom scanner using the scan parameters: 120 kV; 167 mA, slice
thickness, 1mm; matrix size, 512*512, and pixel size, 0.40625. 66,
40 and 40 images are randomly split as training, validating and
testing dataset. 5 different proportions of strongly-annotated data
are tested. For both datasets, we measure the performance by the
Dice score of target segmentation structure between the estimated
and true label maps.

We use the Adam optimizer~\cite{ref21} to train all the models. The
initial learning rate is set to 0.0001 and is reduced by a factor of
0.8 if the mean validation Dice score doesn't increase in 5 epochs.
The training is stopped if the score doesn't increase by 20 epochs.
Dropout~\cite{ref20} is used to the output of each convolution stage
to avoid over-fitting. During the training, we use a mini-batch of 4
images and if the validation Dice score goes up, we evaluate the
model on the testing dataset. The best testing Dice score is
reported as the final result. We perform data augmentation through
random horizontal and vertical flipping, adding Gaussian noise and
randomly cropping the image to a 128*128 patch centered around the
target structure. All images are normalized by subtracting the mean
and dividing by the standard deviation of the training data.

\begin{table}
    \centering
    \caption{Mean of test Dice score (\%).}\label{tab1}
    \begin{tabular}{|c|c|c|c|c|c|}
        \hline
        \multicolumn{6}{|c|}{\textbf{Lung Nodule Segmentation}} \\
        \hline
        \multirow{2}{*}{\textbf{Methods}} & \multicolumn{5}{c|}{\textbf{strong-weak data split(160 in total)}} \\
        \cline{2-6}
        & \textbf{160-0} & \textbf{120-40} & \textbf{100-60} & \textbf{80-80} & \textbf{60-100} \\
        \hline
        \textbf{U-Net} & 84.04$\pm$0.40 & 82.25$\pm$0.39 & 81.85$\pm$0.31 & 80.51$\pm$0.59 & 80.18$\pm$0.97 \\
        \textbf{U-Net+Unary sSE} & 84.01$\pm$0.11 & 82.15$\pm$1.09 & 81.35$\pm$1.39 & 82.08$\pm$0.94 & 80.58$\pm$0.01 \\
        \textbf{Variant MS-Net\cite{ref12}} & - & 82.75$\pm$1.04 & 82.38$\pm$0.53 & 81.72$\pm$1.19 & 79.80$\pm$1.26 \\
        \textbf{MSDN-} & \textbf{84.90$\pm$0.60} & 82.31$\pm$1.14 & 82.17$\pm$0.51 & 81.02$\pm$1.10 & 80.50$\pm$0.37 \\
        \textbf{MSDN} & - & \textbf{83.58$\pm$1.20} & \textbf{83.56$\pm$0.52} & \textbf{83.01$\pm$0.69} & \textbf{82.37$\pm$0.98} \\
        \hline
        \multicolumn{6}{|c|}{\textbf{Cochlea Segmentation}} \\
        \hline
        \multirow{2}{*}{\textbf{Methods}} & \multicolumn{5}{c|}{\textbf{strong-weak data split(66 in total)}} \\
        \cline{2-6}
        & \textbf{66-0} & \textbf{44-22} & \textbf{33-33} & \textbf{22-44} & \textbf{11-55} \\
        \hline
        \textbf{U-Net} & 88.62$\pm$0.08 & 87.41$\pm$0.16 & 86.55$\pm$0.81 & 85.01$\pm$0.39 & 80.85$\pm$0.42 \\
        \textbf{U-Net+Unary sSE} & \textbf{88.97$\pm$0.12} & 87.70$\pm$0.49 & 85.30$\pm$0.28 & 84.38$\pm$0.03 & 82.23$\pm$0.47 \\
        \textbf{Variant MS-Net\cite{ref12}} & - & 87.54$\pm$0.36 & 86.03$\pm$0.25 & 84.71$\pm$0.53 & 82.60$\pm$1.43 \\
        \textbf{MSDN-} & 88.73$\pm$0.33 & 86.73$\pm$1.02 & 85.68$\pm$0.35 & 85.10$\pm$0.15 & 80.81$\pm$0.47 \\
        \textbf{MSDN} & - & \textbf{87.91$\pm$0.28} & \textbf{87.27$\pm$1.08} & \textbf{87.11$\pm$0.28} & \textbf{85.60$\pm$1.76} \\
        \hline
    \end{tabular}
\end{table}

We compare \textbf{MSDN} with other 4 baselines, as shown in
Table.~\ref{tab1}. All the models are trained following the same
setting described aforementioned. A U-Net with the same number of
convolution layers as the segmentation sub-network of our model is
used. For the \textbf{U-Net+Unary sSE}, Unary sSE module is added
after every convolution stage. For the \textbf{Variant MS-Net}, we
follow the thought of MS-Net~\cite{ref12} and build a multi-stream
network based on U-Net, where all features from the Decoder are
taken into the detection stream \textbf{DU}. We also compare to a
reduced version of our model (\textbf{MSDN-}), where we remove the
\textbf{DU} and only preserve the U-NET and the Binary sSE modules.
Note that the \textbf{U-Net}, \textbf{U-Net+Unary sSE} and
\textbf{MSDN-} are trained only with strongly-annotated data. We run
each experiment repeatly for 3 times and the mean dice score with
95\% confidence interval is listed in Table.~\ref{tab1}.

\begin{figure}[ht]
    \centering
    \includegraphics[width=\textwidth]{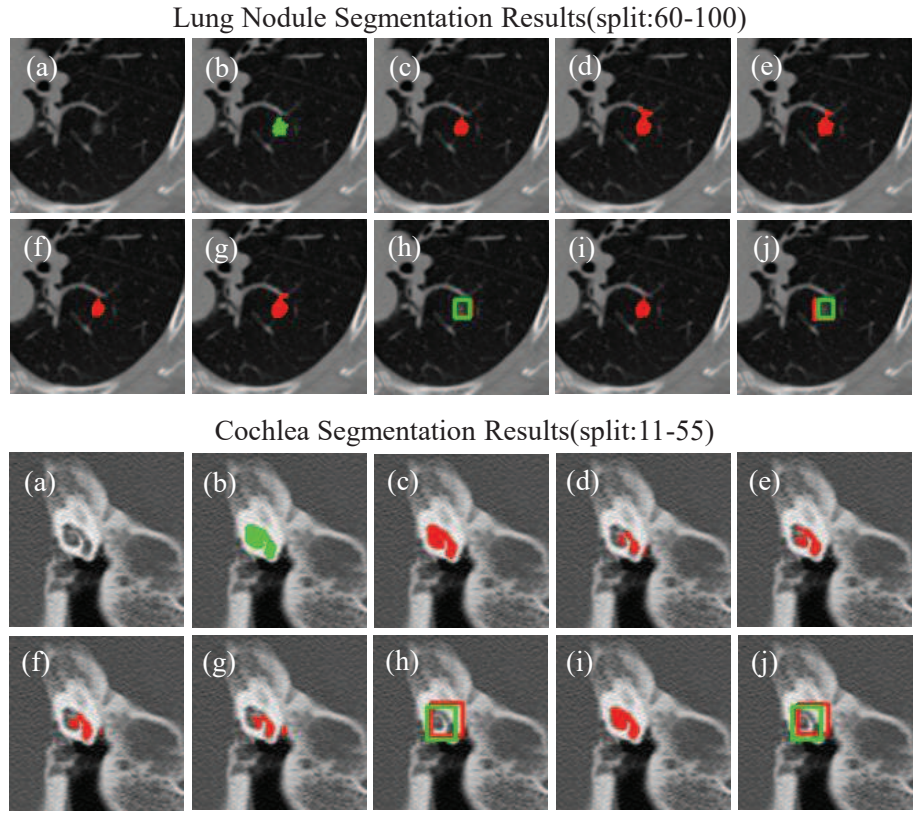}
    \caption{(a) Original image. (b) Ground truth. (c) U-Net trained in full-supervised manner. (d) U-Net trained with only strongly-annotated data. (e) U-Net+Unary sSE. (f) MSDN-. (g),(h) Segmentation and detection results of Variant MS-Net. (i),(j) Segmentation and detection results of MSDN.} \label{fig3}
\end{figure}

From the result, we can see that our model performs better than all
the baselines in the same strong-weak data split. Compared with
models trained in a fully-supervised manner, the performance is
still comparable. When there are few strongly-annotated data for
training, the performances of baselines decrease dramatically (see
the last column). However, the performance of our model still
remains good. The variation of MS-Net improves the results in some
degree, but sometimes the performance is not stable. In contrast,
our model works more stably. The use of sSE module, no matter Unary
or Binary, could improve the training effect, but not as much as our
model, which proves the effectiveness of our design. Some
qualitative results are shown in Fig.~\ref{fig3}.


\section{Conclusion and Future Work}
We propose Mixed-Supervised Dual-Network (MSDN), a novel multi-task
learning architecture for mixed-supervised medical image
segmentation. It is composed of two separate networks for the
detection and segmentation tasks respectively, and a series of sSE
modules as connection between the two networks so that the useful
information of the detection task can be transferred to facilitate
the segmentation task well. We perform experiments on two medical
image datasets and our model outperforms multiple baselines.
Currently, our model can only handle two-task problem. When there
are more than two forms of annotations, our model can not directly
be applied. In the future, we may consider to extend our method to
fit for multi-task scenario \cite{ref22,ref23}.

\subsubsection{Acknowledgement}
This research is supported by China Scholarship Council (CSC).

\end{document}